\documentclass{article}

\usepackage[preprint]{neurips_2026}

\usepackage[utf8]{inputenc}
\usepackage[T1]{fontenc}
\usepackage{amsmath,amssymb,amsfonts}
\usepackage{booktabs}
\usepackage{graphicx}
\usepackage{microtype}
\usepackage{xcolor}
\usepackage{tikz}
\usetikzlibrary{positioning, arrows.meta, calc}
\usepackage{multirow}
\usepackage{array}
\usepackage{float}
\usepackage{placeins}
\usepackage[hidelinks]{hyperref}
\usepackage{url}
\usepackage{enumitem}
\hypersetup{
  pdftitle={Sub-Semantic Image Segmentation},
  pdfauthor={Aviad Cohen Zada and Nadav Orenstein and Shai Avidan and Gal Oren},
  pdfsubject={Preprint},
  pdfkeywords={}
}

\title{Sub-Semantic Image Segmentation}

\author{%
  Aviad Cohen Zada \\
  Tel Aviv University, Israel \\
  \texttt{cohenzada@mail.tau.ac.il} \\
  \And
  Nadav Orenstein \\
  Tel Aviv University, Israel \\
  \texttt{nadavo1@mail.tau.ac.il} \\
  \And
  Shai Avidan \\
  Tel Aviv University, Israel \\
  \texttt{avidan@eng.tau.ac.il} \\
   \AND
  Gal Oren \\
  Stanford University, Technion, USA \\
  \texttt{galoren@stanford.edu} \\
}

\newcommand{\method}{\textsc{Detecture}}
\newcommand{\seg}{\texttt{[SEG]}}
\newcommand{\samthree}{SAM~3}
\newcommand{\qwen}{Qwen3-VL-8B}

\begin{document}
\maketitle

\begin{abstract}
Images can be segmented based on visual cues (i.e., texture segmentation) or into objects (i.e., semantic segmentation). We propose a new category of sub-semantic image segmentation that blurs the line between the two. In sub-semantic image segmentation, language is not used to name whole objects. Instead, it is used to partition an image into stable appearance patterns that can be described by language. To do that, we couple a general-purpose vision-language model to \samthree{}, a promptable segmentation backbone whose native text pathway can ground rich descriptions into masks. Simple coupling fails for a number of reasons that we identify in the paper, and we overcome them by introducing \method{} that resolves three concrete failure modes -- language leakage between texture regions, prompt competition inside the segmentation backbone, and semantic distortion at the language-to-mask interface. Since there is no dataset of sub-semantic image segmentation, we introduce one, termed TextureADE. The new dataset is derived from the ADE20K dataset using a system we designed. We compare \method{} to a number of baselines and find that it achieves the strongest performance on several datasets using different metrics. Code is available at \url{https://github.com/Scientific-Computing-Lab/TextureDetecture}.
\end{abstract}

\begin{figure}[t]
\centering
\includegraphics[width=\linewidth]{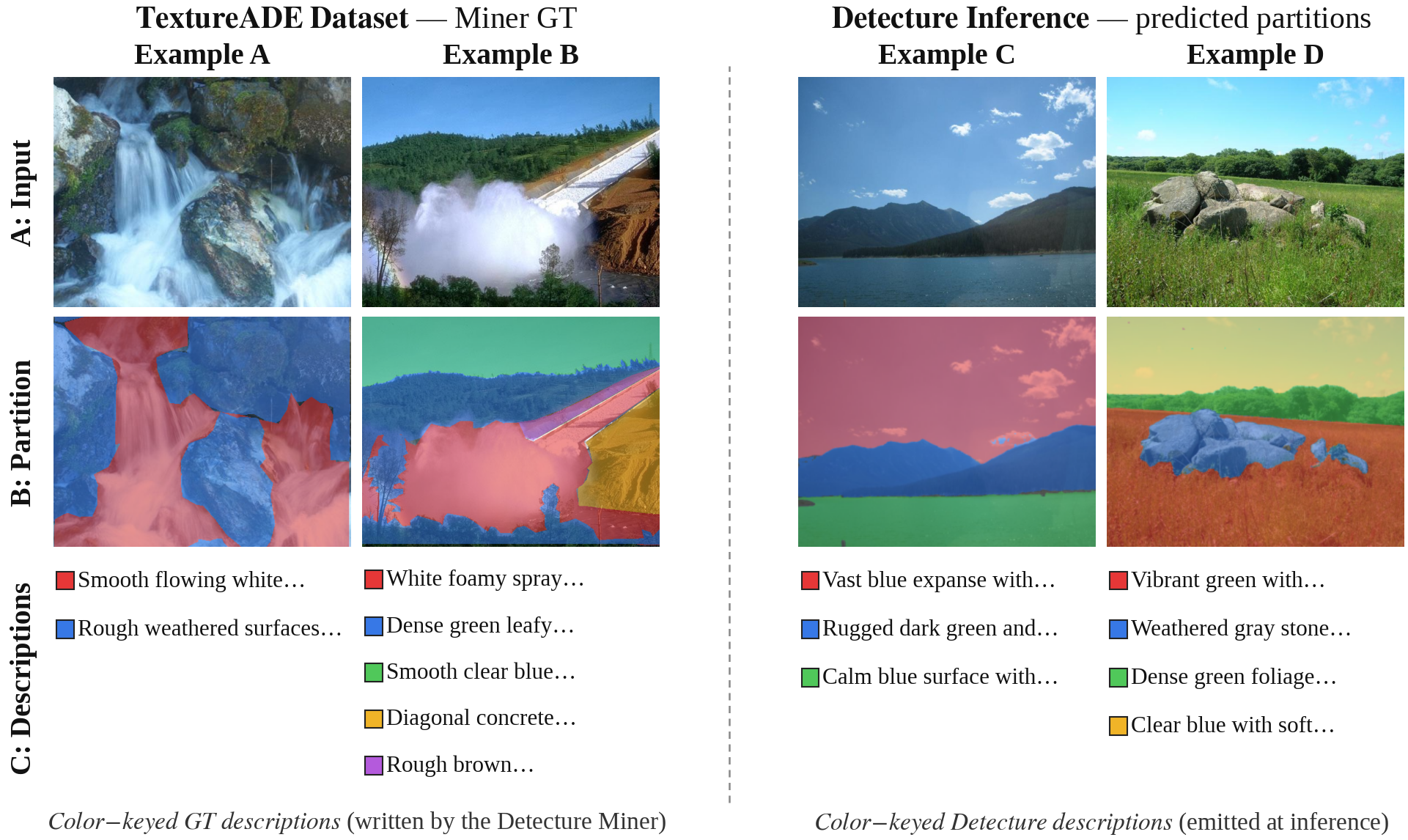}
\caption{\textbf{TextureADE ground truth (left) vs. \method{} inference (right).} Rows display (A) input images, (B) partitions, and (C) color-keyed descriptions across four examples. \textbf{Left:} Ground-truth partitions mined from ADE20K dense masks, with descriptions written by the Detecture Miner (Section~\ref{sec:data_eval}). \textbf{Right:} \method{}'s autonomous predictions on held-out scenes. The inference-time VLM emits descriptions and infers the region count $\hat{K}$ without test-time supervision. (Free-form descriptions are truncated for display; see Appendix~\ref{app:full_descriptions} for complete outputs on novel examples.)}
\label{fig:teaser}
\end{figure}

\section{Introduction}
\label{sec:intro}

Image segmentation spans a spectrum of supervisory cues. At one end, \emph{low-level texture segmentation} relies only on visual statistics: color, gradient, repetition. At the other end, \emph{semantic segmentation} relies on high-level category labels. \emph{Panoptic segmentation} combines the two, partitioning the scene into ``things'' and ``stuff.'' Between purely statistical descriptors and full object categories lies a fourth regime current systems do not address: \emph{sub-semantic} cues -- linguistically stable visual regularities that are smaller than an object category yet still nameable in free-form language. Phrases such as ``thin vertical green strands,'' ``weathered granular gray surface,'' or ``mottled dark background texture'' describe such cues. They cannot be reduced to a fixed class vocabulary, but they are more stable than statistical descriptors because they refer to patterns a human reader can recognize and a vision-language model can produce. This paper argues that sub-semantic cues are the right interface for texture segmentation and provides both a model and a dataset for studying them.

Promptable and open-vocabulary segmentation models~\cite{sam,mask2former,xdecoder,openseed} are strongest when images decompose cleanly into object-like parts. Texture scenes resist that assumption. A boundary between water and sand, gravel and concrete, or fine and coarse fabric need not coincide with object identity, and a region may be better described by material, pattern, scale, and spatial role than by a category label. Texture therefore occupies an intermediate regime between low-level signal and full semantics: local statistics alone are not enough, yet conventional object labels are often too coarse~\citep[e.g.,][]{julesz1981textons,malik1990preattentive}.

That intermediate regime creates an ambiguity that standard segmentation benchmarks often sidestep. A region can be ``grass'' because it belongs to a semantic class, but it can also be ``dense thin green strands'' because that visual regularity is what separates it from soil, pavement, or another plant texture. The latter is not a synonym for the class label. It names an appearance pattern at the scale that matters for mask prediction. In texture-heavy scenes, such patterns may be more stable than the objects that contain them, especially under domain shift, unusual viewpoints, scientific imagery, or material-rich environments where category names are weak predictors of boundaries~\cite{martin2004boundaries,arbelaez2011contour,minc}. Even benchmark families that make non-object regions explicit, such as COCO-Stuff, still describe them through a fixed semantic vocabulary of ``stuff'' classes rather than through free-form appearance partitions~\cite{cocostuff}.

The question, then, is whether language can help in this in-between regime. We argue the useful language is purely sub-semantic: phrases such as ``horizontally stratified red sandstone,'' ``weathered beige stone with vertical striations,'' or ``dense golden-yellow shrubs with mottled pattern'' -- all drawn from our TextureADE corpus -- capture perceptual regularities that are difficult to reduce to classes. Crucially, deciding \emph{which} such phrases to emit is itself a non-trivial reasoning step that weighs spatial context, color, and micro-pattern, lifting the task from passive grounding to autonomous discovery. Such phrases bridge a vision-language model (VLM), which reasons over appearance, and a segmentation model, which grounds prompts into masks. This perspective is adjacent to language-conditioned segmentation~\cite{refexpseg,phrasecut,cmpc,clipseg,cris}, but is harder: the model must first decide \emph{which} descriptions are worth emitting.

We instantiate this idea in \method, an end-to-end inference architecture that, given an image, autonomously generates both a set of text descriptions for the underlying textures and their corresponding segmentation masks -- with no manual, image-specific prompting, no region proposals, and no external $K$-supervision at inference time. The design combines two pretrained components with complementary roles. \qwen~\cite{qwen3vl} is a general-purpose VLM, so we use it for appearance reasoning and for emitting one description per major texture region. \samthree~\citep{sam3} is a promptable segmentation backbone with a native text pathway, so we use it as the mask decoder. \samthree{} on its own has no intrinsic reasoning over \emph{what} to segment -- it is a passive, user-driven tool -- and coupling it with \qwen{} as the cognitive engine is what turns the pair into an autonomous system. The two are connected through dedicated \seg{} hidden states (a special grounding token reserved for mask prediction) that flow through a learned Bridge into \samthree{}'s native text space, where \samthree{}'s own frozen resizer produces decoder prompts; a batch-multiplexed pass yields one mask per texture plus a dustbin, and a Winner-Takes-All assignment turns the overlapping scores into a single partition.

This architecture is not obtained by naively attaching a VLM to SAM. Several standard assumptions behind \seg-token grounding break down when the system is pushed from single-target referring segmentation to autonomous multi-texture partitioning: language states leak across regions, SAM prompt slots compete for pixels, joint training collapses count and emission behavior, and compressed projectors distort semantics. \method{} resolves each failure with a matched mechanism rather than additional capacity. Crucially, by relieving the grounding token from strict linguistic constraints, it allows the backbones to develop a shared visual-spatial dialect. Section~\ref{sec:method} introduces the four ingredients that turn the coupling into a working sub-semantic partition system, and Appendix~\ref{app:pathology_details} reports the diagnostic measurements that motivate each one.

The second contribution is data. Sub-semantic texture segmentation has remained narrow because the supervision needed to study it does not exist at scale. Existing real-world texture benchmarks (e.g., RWTD~\cite{texturesam}, STLD~\cite{texturesam}) contain at most a single texture transition per image; material recognition resources (DTD, MINC, OpenSurfaces) are organized around fixed category labels rather than free-form sub-semantic descriptions~\cite{dtd,minc,opensurfaces,densematerialseg}. To our knowledge, no prior dataset evaluates blind partitioning of multiple texture transitions in realistic scenes. We close this gap with our Detecture Miner, which serves two distinct roles. First, it mines the natural ADE20K validation split to create TextureADE, an in-domain held-out benchmark of real images exhibiting multiple texture transitions. Second, it acts as a strict suitability filter over the synthetically augmented training corpus of TextureSAM~\cite{texturesam}, distilling a massive pool of candidates into a curated set of ${\sim}14{,}000$ multi-texture samples. This intentional asymmetry--training on filtered augmented data and evaluating on natural imagery--is detailed in Section~\ref{sec:data_eval}. Figure~\ref{fig:teaser} previews both halves: the left two columns show TextureADE's ground-truth structure (Examples A, B), and the right two columns show \method{}'s autonomous partitions on held-out samples (Examples C, D). It is important to emphasize the distinct roles of the model and the mining pipeline. The Detecture Miner is not a texture segmentation system; rather, it is a strict data-curation and filtering pipeline that distills large-scale dense-mask corpora into optimal multi-texture training samples. Conversely, \method{} is an autonomous inference model that discovers and segments textures in novel images without relying on pre-existing dense masks.

\begin{figure}[htb]
\centering
\includegraphics[width=\linewidth]{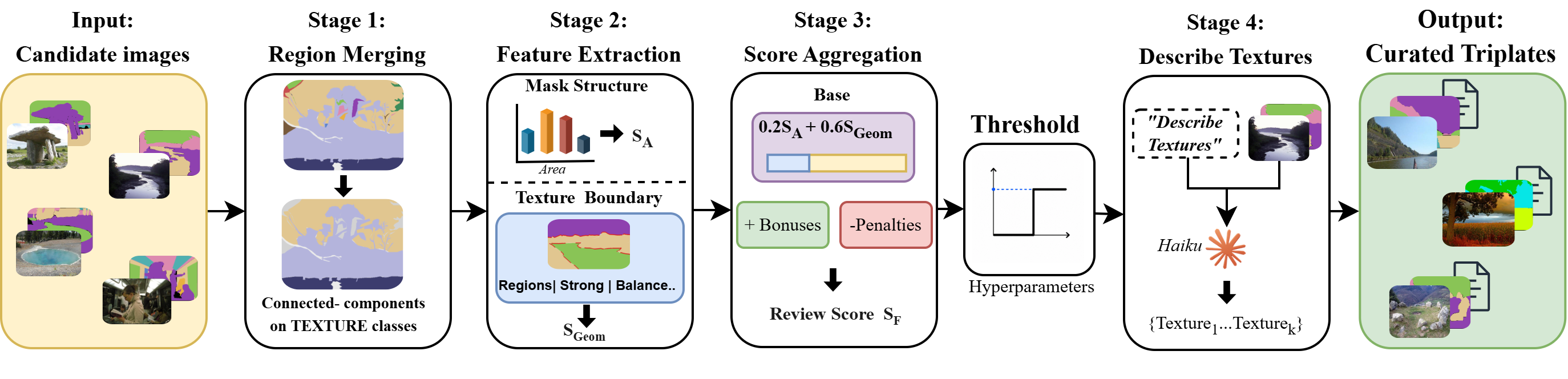}
\caption{\textbf{Detecture Miner data-curation pipeline.} Visual overview of the four stages described in the text; closed-form formulas in Appendix~\ref{app:constructing_textureade}.}
\label{fig:miner_pipeline}
\end{figure}

Figure~\ref{fig:miner_pipeline} summarizes the data-curation pipeline that produces TextureADE and the training corpus, and Figure~\ref{fig:qualitative_results} previews the kind of outputs the paper treats as scientifically meaningful: texture phrases that correspond to perceptually coherent partitions rather than only to category labels.

\textbf{Contributions.} In summary, our main contributions are:
\begin{itemize}[leftmargin=1.2em, labelsep=0.45em, itemsep=0.35em, topsep=0.30em]
  \item \textbf{Sub-semantic image segmentation as a problem class.} We formalize sub-semantic segmentation as the partition of an image into regions described by linguistically stable phrases, bridging the gap between low-level visual cues and semantic object segmentation.

  \item \textbf{Detecture Miner and TextureADE.} We introduce the Detecture Miner, a geometry-first scoring pipeline with a dual role. First, it acts as a strict filter over the augmented corpus of TextureSAM~\cite{texturesam} to yield a curated training set of ${\sim}14{,}000$ multi-texture samples. Second, it mines the natural ADE20K validation split to produce TextureADE, an in-domain held-out benchmark of real images. Unlike existing real-world benchmarks (e.g., RWTD, STLD~\cite{texturesam}), TextureADE is the first to evaluate blind partitioning of multiple texture transitions in realistic scenes.

  \item \textbf{State-of-the-art sub-semantic segmentation.} We propose \method, an end-to-end VLM-to-segmentation architecture specifically designed to ground textual guidance into precise texture masks. Our design prevents representation collapse through isolated token extraction and employs a Shifted-Zero exponential loss on the \seg{} token. Relieving the grounding state from strict language-modeling pressure lets the backbones develop a shared visual-spatial dialect rather than relying on pure text, ensuring accurate geometric guidance without degrading the VLM's open-vocabulary capabilities.
\end{itemize}

For readability, we defer the fuller positioning against neighboring texture, grounding, and promptable-segmentation literatures to Appendix~\ref{app:related_work}, and keep the main paper focused on the scientific question, the new data regime, and the comparative evidence.

\section{Method}
\label{sec:method}

\subsection{Overview}

\begin{figure}[htb]
\centering
\includegraphics[width=0.98\linewidth]{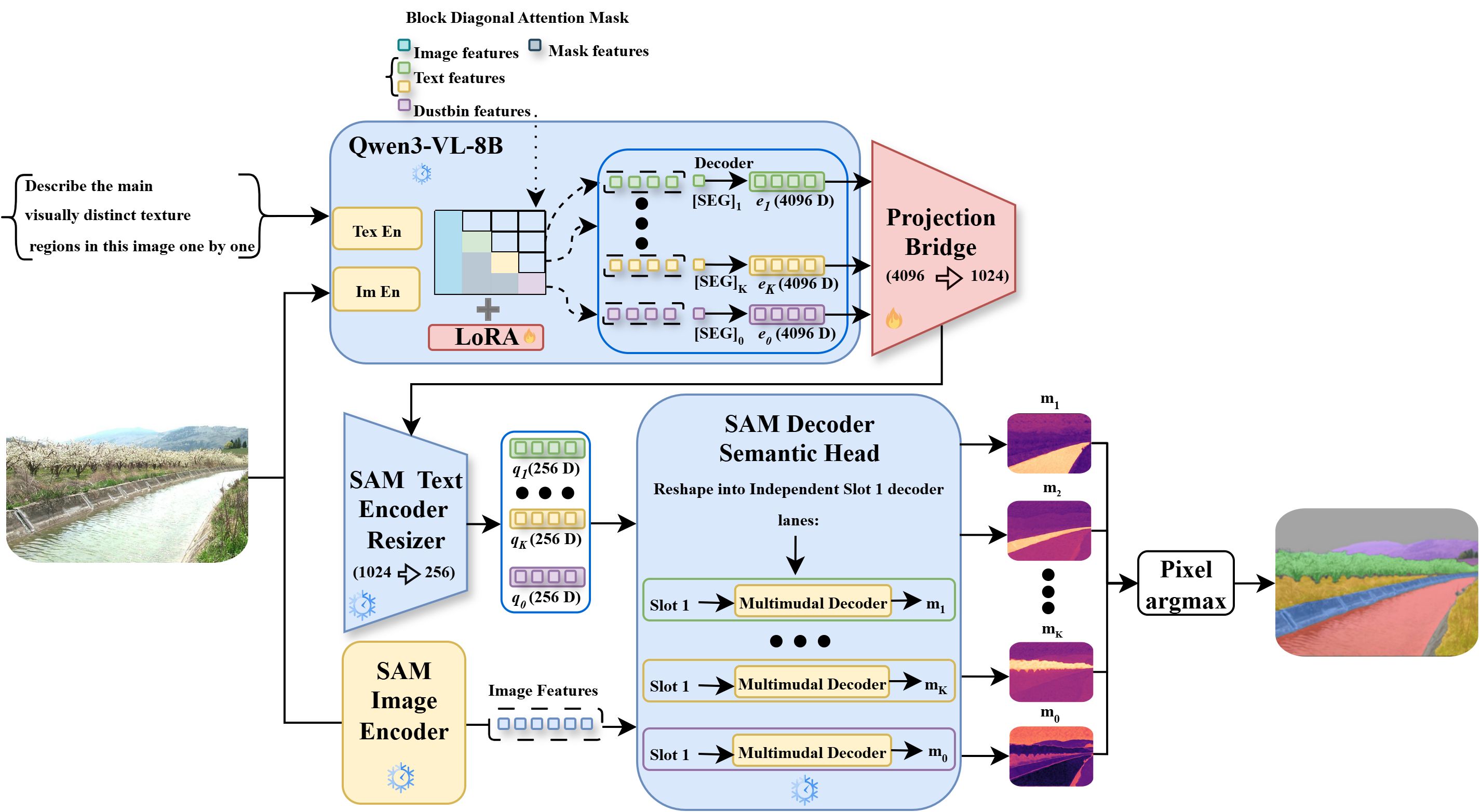}
\caption{\textbf{\method{} inference architecture.} Given an input image and an open-range instruction, \qwen{} emits texture descriptions and \seg{} states under block-diagonal extraction. The Bridge maps each 4096-dimensional state into \samthree{}'s native 1024-dimensional text width. \samthree{}'s frozen resizer then produces 256-dimensional decoder prompts. Batch-multiplexed SAM decoding gives each texture query its own independent Slot~1 before a Winner-Takes-All pixelwise assignment forms the final partition.}
\label{fig:architecture}
\end{figure}

\textbf{Input.} An RGB image $I$. \textbf{Output.} (i) A set of free-form texture descriptions $\{t_k\}_{k=1}^{\hat{K}}$, and (ii) a non-overlapping label map $Y\in\{0,\ldots,\hat{K}\}^{H\times W}$ with label 0 denoting a learned dustbin channel for pixels outside the described regions. \method{} produces both jointly in a single inference pass: \qwen{} names candidate texture regions, the Bridge maps each grounding state into \samthree{}'s native text width, and a Winner-Takes-All assignment converts the per-texture masks into a partition. Figure~\ref{fig:architecture} illustrates the full data path.

\subsection{Notation and settings}

Let $I$ be an input image and $\{t_k\}_{k=1}^{\hat{K}}$ the predicted texture descriptions, each of which is a short free-form phrase that names material or appearance and provides enough spatial contrast to ground a region. The descriptions need not be class names. The associated partition map $Y\in\{0,\ldots,\hat{K}\}^{H\times W}$ assigns every pixel either to one described texture ($1{\le}k{\le}\hat{K}$) or to a learned dustbin (label 0; equivalently, the $(\hat{K}+1)$-th competing channel in the softmax). The model receives one of two instructions: an \emph{exact-}$K$ instruction, which fixes the count for controlled evaluation, or an \emph{open-range} instruction, which forces \method{} to infer $\hat{K}$ on its own. We use the exact-$K$ setting only as a controlled diagnostic of texture separation at fixed perceptual granularity (not as a count-prediction probe), and reserve the open-range setting for autonomous region-count inference. Treating \method{} as a partition model -- where all texture logits and the dustbin compete at every pixel -- distinguishes it from independent referring-segmentation calls, which do not define a partition: overlapping masks have no internal arbiter, and a single emitted region can hide a multi-texture failure under Hungarian matching. Training uses (image, partition, description-set) triples distilled by the Detecture Miner from the augmented corpus of TextureSAM~\cite{texturesam}; the optimization objective is a weighted sum of a cross-entropy loss on the assistant-region description tokens (with a Shifted-Zero schedule that protects the \seg{} hidden state from ordinary linguistic pressure) and a Dice~$+$~Focal loss on the partition logits. The exact loss formulation, weighting schedule, and two-stage curriculum (Stage~1: Bridge + mask head; Stage~2: \qwen{}~LoRA on attention $q,v$ projections) are summarized in \emph{Training and partitioning} below and detailed in Appendix~\ref{app:pathology_details}.

\subsection{Architecture}

The architecture follows the same three-step logic as the scientific question: describe candidate texture regions, translate those descriptions into the segmentation backbone's prompt space, and force the resulting masks to compete as a partition. Given an image, \method{} predicts $\hat{K} \leq 6$ texture descriptions, their masks, and a dustbin for pixels outside the described regions. The language-model output follows the structured form
\[
\texttt{TEXTURE\_k: Texture of <description> \seg},
\]
where the final \seg{} token marks the hidden state that should drive mask prediction. Each assistant-region \seg{} hidden state $\mathbf{h}_k \in \mathbb{R}^{4096}$ therefore becomes a grounding vector. A learned dustbin vector is appended, producing $\hat{K}{+}1$ query states. Because the language model and segmentation backbone operate in different hidden widths, a learned Bridge translates each state into the format expected by \samthree{}:
\[
f_B(\mathbf{h})=\mathrm{Dropout}_{0.4}\!\left(\mathrm{GELU}\!\left(\mathrm{LN}(W_B\mathbf{h}+b_B)\right)\right)
\]
This maps each state to $\mathbb{R}^{1024}$. \samthree{}'s frozen text resizer $R:\mathbb{R}^{1024}\to\mathbb{R}^{256}$ then produces decoder prompts. We keep \samthree{}'s image encoder, text resizer, transformer, and mask decoder frozen throughout training so that the learned adaptation occurs primarily at the interface between language and masks.

\subsection{Pathologies and fixes}

Each remaining design choice answers a specific failure along this prediction path. The main paper keeps the logic in view and defers the exact masking rules, loss weights, and quantitative diagnostics to Appendix~\ref{app:pathology_details}.

\textbf{Context leakage.}
Under standard causal decoding, later \seg{} states absorb earlier texture descriptions. We therefore generate the descriptions first and then recompute the assistant sequence under block-diagonal attention, so each texture block can see the shared visual prefix and its own words, but not its neighboring texture blocks.

\textbf{SAM slot competition.}
The next failure comes from the segmentation backbone rather than the language model. When several learned prompts are passed to \samthree{} together, prompt slots compete positionally rather than semantically. Batch multiplexing avoids that route by decoding each texture query as Slot~1 in its own batched lane.

\textbf{Count and emission collapse.}
Uniform language loss encourages the model to stop early, while removing language loss altogether breaks description quality and \seg{} emission. Shifted-Zero keeps normal linguistic pressure where ordinary words matter, removes it from the grounding state itself, and combines with masked-row \seg{} training so emission remains learnable at inference time.

\textbf{Projector drift.}
The final failure appears at the interface between the two frozen backbones. Overfit or bottleneck projectors distort semantics that \samthree{}'s native text pathway already understands. Bridge-to-Native-Space therefore maps \qwen{} states into \samthree{}'s native text width and leaves \samthree{}'s own text resizer frozen.

\subsection{Training and partitioning}

Training mirrors the architecture's division of labor. During training, the model receives the open-range instruction and is optimized via teacher-forcing on the ground-truth descriptions, forcing it to learn autonomous emission and stopping behavior without relying on an external region count. We first stabilize the language-to-mask interface by training only the Bridge, mask head, and dustbin, and only then unlock lightweight LoRA on the language model together with the masked \seg{} rows. \samthree{} remains frozen throughout. Optimization combines mask cross-entropy, Dice, and Shifted-Zero language supervision; the exact schedule, weights, and masking details are given in Appendix~\ref{app:pathology_details}.

At inference, the texture queries and the dustbin compete pixelwise in a Winner-Takes-All assignment, producing a partition rather than a set of independently thresholded masks. Hungarian matching is used only at evaluation time, where unordered predicted regions must be aligned with unordered ground-truth regions.

\section{TextureADE and Evaluation Protocol}
\label{sec:data_eval}

Having established the dataset rationale in Section~\ref{sec:intro}, we designate TextureADE as the primary realistic-scale autonomous-discovery benchmark, with the blind variable-$K$ regime as its defining feature. RWTD and STLD~\cite{texturesam} are reserved as controlled exact-$K{=}2$ ablations that isolate boundary precision under fixed perceptual granularity, with RWTD additionally serving as a cross-domain stress test. Appendix~\ref{app:constructing_textureade} expands on this evidentiary split.

\textbf{Constructing the Data.} The Detecture Miner is the offline data-curation engine behind both TextureADE and the training corpus; the same four-stage pipeline runs unchanged on both. \emph{Stage~1 (Region Merging)} runs 4-connectivity connected-components labeling on per-class TEXTURE pixels and merges the resulting fragments into at most five candidate regions per image, each required to cover at least $1\%$ of image area. \emph{Stage~2 (Feature Extraction)} scores each candidate set along two complementary axes: a mask-structure score $s_A$ (region count, size balance, area-distribution entropy) and a texture-boundary geometry score $s_{\mathrm{geom}}$ (large-region count, strong inter-region boundaries, boundary length normalized by image perimeter, area balance over large regions). \emph{Stage~3 (Score Aggregation)} combines them into a single review score $s = 0.20\,s_A + 0.60\,s_{\mathrm{geom}} + \mathrm{bonuses} - \mathrm{penalties}$, where bonuses reward two-to-four coherent textured regions and penalties downweight object-dominated, mosaic-like, or fragmented scenes; an image is admitted iff $s\geq 65/100$. The geometry-dominant weighting embodies the geometry-first design: acceptance rests on observable mask geometry, not on VLM judgment. \emph{Stage~4 (Language-Second Annotation)} renders an image\,$+$\,mask overlay for each admitted region and queries a frozen VLM annotator with a fixed structured template, tying each description to a proposed region rather than letting the VLM invent regions from the raw image. Run on the CNT-augmented~\cite{cnt} ADE20K \emph{training} split of TextureSAM, the Miner distills ${\sim}10^{5}$ candidates into ${\sim}14{,}000$ training samples; run on the \emph{natural} ADE20K validation split, it accepts $212$ samples that form TextureADE. Closed-form definitions of every component, the bonus and penalty terms, and the annotation template are in Appendix~\ref{app:constructing_textureade}.

\textbf{Two evaluation regimes.}
The benchmark suite isolates two distinct failure modes. RWTD ($253$ natural two-texture images) and STLD ($200$ controlled two-texture samples) are evaluated under a controlled exact-$K{=}2$ \emph{Oracle} protocol that fixes the count by task definition. \method{} and Sa2VA~\cite{sa2va} receive $K{=}2$ via prompt wording; SAM-based baselines, which lack autonomous stopping rules and native multi-partition logic, receive the standard \emph{inverse-mask construction} (predict one mask, take its complement as the second). We frame this construction explicitly as a \emph{generous concession} granted to those baselines: it externally solves the count and partitioning problem they cannot solve themselves, and additionally prevents the oversegmentation they otherwise produce in dense-proposal mode -- where many small candidate regions would tank mIoU through partition fragmentation rather than through textural-boundary error -- so they can be evaluated on boundary quality alone, the regime in which they are designed to compete. TextureADE ($212$ ADE20K-derived samples, $K$ hidden at test time) is the complementary blind variable-$K$ \emph{Autonomous} regime, which tests whether a method can decide $K$ for itself. Among baselines only \samthree{} receives a top-$K_{\mathrm{GT}}$ concession on TextureADE because its proposal route lacks a native stopping rule; that concession also favors the baseline. We compare against four baselines (\samthree, Grounding-\samthree, Sa2VA, TextureSAM) covering native text path, detector-driven proposals, conversational VLM grounding, and prompt-following augmented supervision; per-baseline fairness details are in Appendix~\ref{app:eval_protocol}.

We report two complementary metrics: mean intersection-over-union (mIoU) under Hungarian matching, and Adjusted Rand index (ARI) over the full partition, which is sensitive to clustering structure rather than only foreground overlap. The two can disagree -- on STLD \method{} leads mIoU while Sa2VA and TextureSAM have higher ARI -- so we report metric-specific conclusions rather than collapsing them.

\section{Experiments}
\label{sec:experiments}

\textbf{Training and inference setup.}
The final checkpoint follows the two-stage schedule of Section~\ref{sec:method} on the Detecture-Miner-curated training corpus. Adaptation is small: LoRA on \qwen{}'s attention $q,v$ projections, Bridge dropout 0.4, $K\!\leq\!6$. At inference, the model emits descriptions and \seg{} tokens, recomputes isolated \seg{} states under block-diagonal attention, and decodes the query batch through frozen \samthree{}.

\textbf{Main results.}
Table~\ref{tab:main_results} compares \method{} against four baselines. \samthree{}~\citep{sam3} (vanilla) tests the native promptable backbone. \emph{Grounding-\samthree{}}~\cite{groundingdino} serves as the community gold standard for robust ``detector-to-segmenter'' pipelines. \emph{Sa2VA}~\cite{sa2va} is the state-of-the-art in reasoning-guided segmentation; while it excels at complex reasoning and single-target extraction, it probes whether conversational referring can substitute for competitive multi-region partitioning. Finally, \emph{TextureSAM}~\cite{texturesam} tests an alternative supervision route. Crucially, while TextureSAM was trained on the entire CNT-augmented ADE20K corpus, \method{} was trained on a strictly curated, non-random subset comprising only ${\sim}14\%$ of that exact same data, yet demonstrates superior performance across nearly all metrics. Per-baseline instantiation and concession rules are in Appendix~\ref{app:eval_protocol}.

\begin{table}[htb]
\centering
\small
\setlength{\tabcolsep}{4pt}
\begin{tabular}{@{}llccccc@{}}
\toprule
Route & Metric & SAM~3 & Grounding-SAM~3 & Sa2VA & TextureSAM & \textsc{Detecture} \\
\midrule
\multirow{2}{*}{RWTD}
  & mIoU & 0.6337 & 0.4640 & 0.3561 & 0.4684 & \textbf{0.8053} \\
  & ARI & 0.4427 & 0.1959 & 0.5593 & 0.6163 & \textbf{0.6782} \\
\midrule
\multirow{2}{*}{STLD}
  & mIoU & 0.5042 & 0.4489 & 0.3739 & 0.4677 & \textbf{0.7069} \\
  & ARI & 0.1378 & 0.0539 & \textbf{0.7140} & 0.6849 & 0.5676 \\
\midrule
\multirow{2}{*}{TextureADE}
  & mIoU & 0.3194 & 0.4518 & 0.7141 & 0.4798 & \textbf{0.7328} \\
  & ARI & 0.3219 & 0.4794 & 0.7011 & 0.3566 & \textbf{0.7047} \\
\bottomrule
\end{tabular}

\caption{\textbf{Cross-dataset texture segmentation.} RWTD and STLD: controlled exact-$K{=}2$ routes (cross-domain, matched-domain). TextureADE: in-domain held-out blind variable-$K$. \method{} leads mIoU on all three and ARI on RWTD and TextureADE.}
\label{tab:main_results}
\end{table}

\begin{figure}[!htb]
\centering
\includegraphics[width=0.78\linewidth]{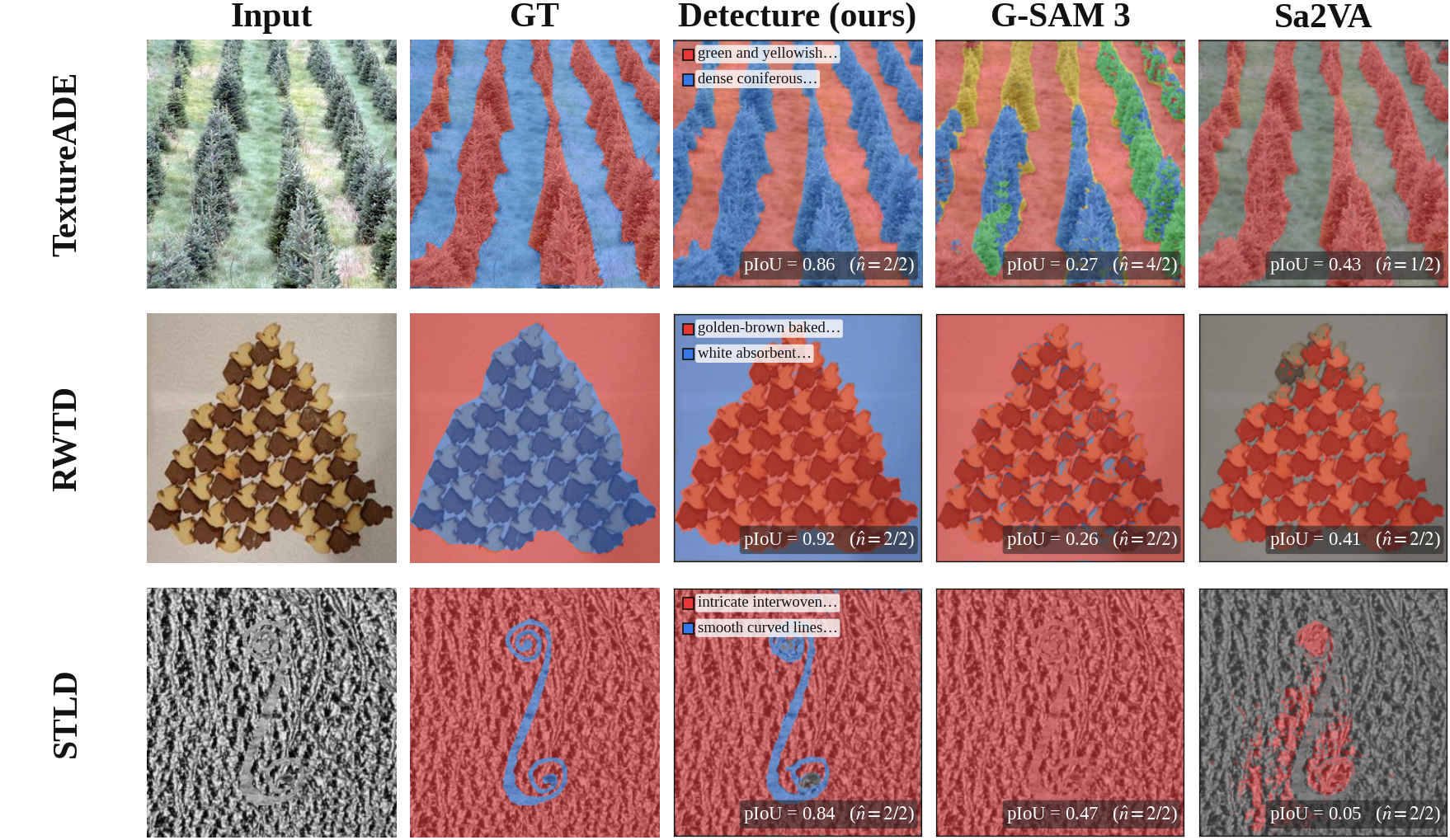}
\caption{\textbf{Qualitative results across the three routes.} \method{} is the only model emitting structured texture descriptions alongside masks (color-keyed strip per \method{} cell; truncated). Vanilla \samthree{} omitted. Sa2VA's TextureADE row exemplifies the autonomous-stopping failure: clean mask (mIoU~$=0.86$) but $\hat{n}{=}1$ vs.\ $K_{\mathrm{GT}}{=}2$ halves pIoU to $0.43$.}
\label{fig:qualitative_results}
\end{figure}

For qualitative analysis, we report the per-image Hungarian-matched IoU (pIoU) for individual examples. Figure~\ref{fig:qualitative_results} shows the input, ground truth, \method{}'s predicted partition (with its emitted phrase strip), and the two strongest baselines (Grounding-\samthree{} and Sa2VA). The pIoU and predicted region counts are overlaid on each cell. Untruncated descriptions for fifteen novel samples are provided in Appendix~\ref{app:full_descriptions}.

\FloatBarrier

\textbf{Ablation summary.}
The full ablation trail is in Appendix~\ref{app:ablation}: teacher-forced rows diagnose the geometry path, live rows diagnose description and \seg{} emission, and the zero-shot \samthree{} diagnostic clarifies why native-width alignment matters. To rule out semantic-object leakage we additionally perform a strict head-noun omission ablation, deterministically stripping the leading anchor before the \seg{} hidden state is computed. Because the Shifted-Zero schedule frees the \seg{} state from ordinary language-modeling pressure, it functions as a visual-spatial dialect that grounds textural features rather than class labels: \method{} retains SOTA mIoU on RWTD/STLD/TextureADE under strict anchor omission (Table~\ref{tab:semantic_ablation}; visual evidence in Appendix~\ref{app:ablation}).

\begin{table}[htb]
\centering
\small
\setlength{\tabcolsep}{4pt}
\newcommand{\cmark}{\textcolor{black!85}{\checkmark}}
\newcommand{\xmark}{\textcolor{black!50}{$\times$}}
\begin{tabular}{@{}llccccr@{}}
\toprule
\multirow{2}{*}{Route}
  & \multirow{2}{*}{Metric}
  & \multirow{2}{*}{\shortstack[c]{Strongest baseline\\(value)}}
  & \multicolumn{2}{c}{\method{}}
  & \multirow{2}{*}{$\Delta$}
  & \multirow{2}{*}{\shortstack[c]{Still\\best?}} \\
\cmidrule(lr){4-5}
  & & & Before & After & & \\
\midrule
\multirow{2}{*}{RWTD}
  & mIoU & SAM~3       (0.6337) & 0.8162 & \textbf{0.8053} & $-0.0109$ & \cmark \\
  & ARI  & TextureSAM  (0.6163) & 0.6895 & \textbf{0.6782} & $-0.0113$ & \cmark \\
\midrule
\multirow{2}{*}{STLD}
  & mIoU & SAM~3       (0.5042) & 0.7441 & \textbf{0.7069} & $-0.0372$ & \cmark \\
  & ARI  & Sa2VA       (0.7140) & 0.6062 &        0.5676   & $-0.0386$ & \xmark \\
\midrule
\multirow{2}{*}{TextureADE}
  & mIoU & Sa2VA       (0.7141) & 0.7419 & \textbf{0.7328} & $-0.0091$ & \cmark \\
  & ARI  & Sa2VA       (0.7011) & 0.7138 & \textbf{0.7047} & $-0.0091$ & \cmark \\
\bottomrule
\end{tabular}

\caption{\textbf{Semantic Anchor Ablation.} \method{} performance with the head noun \emph{included} (Before) vs.\ \emph{deterministically stripped} (After). Under strict stripping, \method{} retains SOTA mIoU on every route and SOTA ARI on RWTD and TextureADE; \cmark/\xmark{} marks whether the After value is still best across the five-method comparison. Visual evidence: Appendix~\ref{app:ablation}, Figure~\ref{fig:semantic_anchor_ablation}.}
\label{tab:semantic_ablation}
\end{table}

\FloatBarrier
\section{Conclusion}
\label{sec:conclusion}

\method{} shows that language helps texture segmentation when it operates as a sub-semantic coordination layer rather than as an object-label interface. What makes the system work is not the bare coupling of a VLM and \samthree{}, but a set of constraints that preserve the right representation at each stage: isolate grounding states, prevent prompt-slot competition, preserve emission behavior, and bridge into the decoder's native semantic space. TextureADE supplies the missing data regime that lets this question be studied at realistic scale rather than only on tiny real-world benchmarks or synthetic alternatives. Across controlled exact-$K$ two-texture separation on RWTD and STLD and blind variable-$K$ partitioning on TextureADE, this combination leads mIoU on all three routes, supporting a precise conclusion: sub-semantic language grounding is an effective interface for multi-texture partitioning under controlled evaluation regimes.

\section{Limitations}
\label{sec:limitations}

\textbf{Generation cost and backbone dependence.}
\method{} is slower than prompt-only segmentation because it first generates texture descriptions and then recomputes isolated \seg{} states, so latency grows with the number and length of emitted regions. The system also depends on \samthree{}'s frozen text prior: this helps generalization, but it is also a ceiling. The strong zero-shot \samthree{} diagnostic suggests that better bridging can help only up to the point allowed by the backbone's own concept coverage, so domains with visually meaningful but linguistically unfamiliar textures may require backbone adaptation rather than only a better Bridge.

\textbf{Evaluation scope.}
The current benchmark universe remains narrow because the field itself lacks large real-scene texture datasets. RWTD is small, STLD is a controlled exact-$K$ two-texture separation route, and TextureADE is ADE20K-derived. We treat that last fact as both strength and limitation: TextureADE is the in-domain but held-out realistic-scale route this problem has been missing, but it still inherits ADE20K and miner-induced biases. RWTD is therefore the stronger cross-domain stress test, while TextureADE carries the complementary burden of testing blind variable-$K$ partitioning at realistic scale. RWTD and STLD are intentionally not blind count-prediction tests; they isolate texture-boundary quality once the perceptual granularity is fixed. All three routes remain texture-focused rather than general open-vocabulary scene partitioning.

\textbf{Supervision ambiguity.}
Detecture Miner is geometry first and language second, which reduces hallucinated regions but does not make the resulting descriptions uniquely correct. Many ADE20K-derived images admit several valid texture decompositions depending on scale and intent, and the language labels are training supervision rather than claims about the only correct human description of a region.

\section*{Acknowledgments}
This work was supported by the Pazy Foundation.

\clearpage
\bibliographystyle{plainnat}
\bibliography{references}

\appendix
\section{Related Work Context}
\label{app:related_work}

The main paper defers the fuller literature positioning to this appendix so that the core text can stay focused on the scientific question, the new data regime, and the comparative evidence. Three neighboring literatures shape the present question. One asks how textures and materials should be represented. A second asks how language can serve as an interface for dense prediction. A third asks how modern promptable segmentation systems convert text or proposals into masks. Our setting lies at the intersection: we want language to help a model discover texture regions for itself, not merely follow a supplied target.

\textbf{Texture and material segmentation.}
Classical work on texture perception emphasizes textons and preattentive texture differences~\cite{julesz1981textons,malik1990preattentive}, while descriptor-based and statistical models such as local binary patterns and filter-bank statistics formalized texture as a discriminative appearance signal~\cite{ojala2002lbp,varma2005texture}. Early cue-integration formulations and later boundary-based segmentation work likewise showed that texture cues are central to separating regions in natural images~\cite{textonscontours,martin2004boundaries,arbelaez2011contour}. Recognition datasets such as DTD~\cite{dtd}, KTH-TIPS~\cite{kth}, and MINC~\cite{minc}, together with material-recognition work in the wild~\cite{sharan2013material}, tell us what texture and material categories look like, while OpenSurfaces, per-pixel material-context work, and newer dense material segmentation datasets bring surface labels into real scenes~\cite{opensurfaces,visualtraits,densematerialseg}. None of these resources, however, asks a model to autonomously partition a full image into blind texture regions. TextureSAM~\cite{texturesam} moves closer to our setting by directly addressing the shape bias of SAM-family models with texture-augmented supervision built from ADE20K~\cite{ade20k} via Compositional Neural Textures~\cite{cnt}. That design choice is already informative: large real-scene texture supervision is scarce, and ADE20K is one of the few dense-mask resources that can be repurposed into texture regions at meaningful scale. TextureSAM showed that SAM-style models can be shifted toward texture boundaries, but the setup remains prompt driven: the number and identity of target regions are supplied externally. \method{} instead studies the harder upstream problem of discovering and expressing those regions before mask decoding. We reuse TextureSAM's central insight that texture supervision matters, but move the question from ``can SAM follow a texture prompt?'' to ``can a model decide and ground the set of texture prompts itself?''

\textbf{Language-guided dense grounding.}
A second line of work treats language as a handle for dense prediction. Early referring-expression segmentation asked a model to return the mask named by a supplied expression~\cite{refexpseg}, and PhraseCut~\cite{phrasecut} broadened that regime to open-domain phrases in natural images. CMPC~\cite{cmpc}, CLIPSeg~\cite{clipseg}, and CRIS~\cite{cris} strengthened language-conditioned mask prediction before the multimodal-LLM wave. LISA~\cite{lisa} introduced a learnable \seg{} token whose hidden state conditions a SAM mask decoder; LISA++~\cite{lisapp} extended this line of work. GSVA~\cite{gsva} generalized the idea to multiple grounding tokens and explicit rejection. Recent models like Sa2VA~\cite{sa2va} achieve state-of-the-art results in reasoning-guided segmentation by marrying a LLaVA-style VLM with a segmentation backbone via specialized grounding tokens. However, these systems are fundamentally optimized for single-target referring tasks where the user specifies the target. When tasked with autonomous multi-region discovery, they require independent, sequential prompts. This independent extraction not only exposes the system to causal-token contamination and prompt-slot competition, but crucially, it does not define a partition: independently thresholded masks can overlap, leave gaps, or describe inconsistent granularities. A true texture partition requires the targets to explicitly compete for pixels.

\textbf{Promptable and open-vocabulary segmentation backbones.}
A third line of work concerns the masking backbones themselves. SAM~\cite{sam}, SAM~2~\cite{sam2}, and \samthree~\citep{sam3} provide strong prompt-conditioned segmentation systems; \samthree{} is especially relevant here because it adds a native text pathway, making it possible to test whether rich appearance descriptions can be grounded directly. Grounding-DINO~\cite{groundingdino} plus SAM is a common detector-to-mask pipeline in which open-set detections propose regions before masking. Datasets such as COCO-Stuff~\cite{cocostuff} make non-object regions explicit at scale, and panoptic or open-vocabulary methods such as Panoptic FPN~\cite{panopticfpn}, Mask2Former~\cite{mask2former}, X-Decoder~\cite{xdecoder}, and OpenSeeD~\cite{openseed} predict sets of masks across thing and stuff categories. Yet these systems still usually operate against fixed or extensible semantic vocabularies rather than free-form texture descriptions. \method{} differs in two ways: its intermediate representation is appearance language rather than category labels, and its goal is a partition in which multiple descriptions compete for ownership of each pixel.

\section{Additional Method and Diagnostic Details}
\label{app:pathology_details}

\subsection{Early VLM-to-SAM Prototype}

This appendix records the intermediate designs and implementation details that support the main scientific claim. We begin with the earliest prototype, because it clarifies which parts of the final system are conceptual advances rather than cosmetic refinements. The original draft described an early, non-final prototype that used Qwen2.5-VL, an earlier version of the vision-language backbone, together with marker tokens such as \texttt{<START\_SEG\_A>} and \texttt{<END\_SEG\_A>} to extract hidden states around generated texture descriptions. Those states were projected through a multi-token MLP into a SAM-family decoder. The prototype was useful as a feasibility test: VLM hidden states carried enough appearance information to guide a segmentation backend, and competitive pixel assignment was more coherent than independent mask thresholding. It was not, however, the final architecture. The model often produced generic descriptions such as ``rough surface'' where the mask decoder needed a more discriminative phrase, and its marker-based extraction did not solve multi-texture leakage. The final system replaces this design with \qwen{}, explicit \seg{} rows, block-diagonal extraction, and Bridge-to-Native-Space.

\subsection{Block-Diagonal Extraction}

The first implementation detail concerns how to keep one texture description from contaminating the next. Let the assistant output contain $K$ texture blocks $\mathcal{G}_1,\ldots,\mathcal{G}_K$, each ending in an assistant-region \seg{} token, and let $\mathcal{P}$ denote the shared prefix containing image tokens, system prompt, and user instruction. In the extraction pass, each token in $\mathcal{G}_k$ may attend to $\mathcal{P}$ and to earlier tokens in $\mathcal{G}_k$, but not to any token in $\mathcal{G}_{\ell\neq k}$. This preserves autoregressive ordering within a texture description while preventing later \seg{} states from absorbing earlier texture semantics. The main paper reports the resulting diagnostic cosine shift from 0.16 under isolated extraction to 0.74 under shared context.

\subsection{SAM Slot Bias and Batch Multiplexing}

The second detail concerns the segmentation backbone rather than the language model. A slot diagnostic measured how SAM allocates pixels when it receives more than one learned prompt at once. On RWTD, the first slot received 90.5\% of pixels and the second 0.0\%, and the pattern persisted after texture-order swap. The final implementation avoids that multi-prompt route by reshaping query states from $B\times K\times D$ to $(BK)\times 1\times D$, duplicating the indexed image embedding, and running a single batched SAM call. This preserves GPU batching while ensuring that every texture query is decoded as Slot~1 in its own lane.

\subsection{Shifted-Zero Indexing}

The third detail concerns the language objective itself. The language loss is applied only over assistant-region tokens. The intended behavior is threefold: preserve ordinary linguistic pressure on description tokens, make emitting \seg{} likely at inference time, and avoid forcing the \seg{} hidden state to predict ordinary continuation text. In implementation terms, the token-position weight is full at assistant \seg{} positions, zero at the position immediately after each assistant \seg, and exponentially restored away from the boundary. User-prompt format examples containing the literal \seg{} token are filtered out before computing distances, so template tokens are not treated as grounding outputs.

\subsection{Bridge-to-Native-Space}

The fourth detail concerns the interface between the two pretrained backbones. The Bridge is a single native-width mapping
\[
f_B(\mathbf{h})=\mathrm{Dropout}_{0.4}\!\left(\mathrm{GELU}\!\left(\mathrm{LN}(W_B\mathbf{h}+b_B)\right)\right),
\]
where $\mathbf{h}\in\mathbb{R}^{4096}$ and $f_B(\mathbf{h})\in\mathbb{R}^{1024}$. The subsequent $1024{\to}256$ mapping is \samthree{}'s frozen text resizer. The original draft contrasted this design with earlier $4096{\to}512{\to}256$ and high-capacity projector variants. The key diagnostic was that \samthree{}'s native text path could reach 0.928 RWTD mIoU with hand-written object-aware prompts, so the final projector is designed to preserve the segmentation backbone's text geometry rather than learn a replacement low-dimensional semantic map.

\subsection{Training Curriculum}

The final curriculum has two stages, reflecting the same logic as the main paper: first stabilize the language-to-mask interface, then lightly adapt the language model. Stage 1 trains only the Bridge, mask head, and dustbin embedding for epochs 1--8. Stage 2 unlocks \qwen{} LoRA on the attention query and value projections, masked-row \seg{} input/output rows, and a decayed Bridge learning rate for epochs 9--30.

The trainable budget stays concentrated at the interface rather than in the frozen backbones: the Bridge contributes about 4.2M parameters, \qwen{} LoRA about 3.8M, and the masked \seg{} rows add only 16{,}384 effective parameters, for roughly 8.2M trainable parameters in total.

\textbf{Training-time prompt and supervision.}
Both training stages use the \emph{open-range} instruction (the instruction \method{} uses for the autonomous-discovery regime on TextureADE; the exact-$K$ probe used on RWTD/STLD is a controlled diagnostic and is not part of training); the count is never given as a prompt argument. Supervision is teacher-forcing on the ground-truth assistant region: at every step the previous tokens are the ground-truth description tokens, and the cross-entropy loss is applied on the next-token prediction with the per-position weighting described in \emph{Shifted-Zero Indexing} above. Because the GT itself terminates after the correct number of textures, the model learns to terminate after the correct number of textures purely from imitation -- there is no auxiliary count head, no $K$-conditioning input, and no explicit stop signal.

\textbf{Loss weights and schedule.}
The total objective is $\mathcal{L} = \mathcal{L}_{\text{CE}}^{\text{mask}} + 3.0\cdot\mathcal{L}_{\text{Dice}} + 0.1\cdot\mathcal{L}_{\text{LM}}^{\text{SZ}}$. The Shifted-Zero exponential cliff uses $\alpha = 2.0$, so the weight at distance 1 from an assistant \seg{} is 0.865, at distance 2 is 0.982, and at distance ${\ge}3$ is essentially 1.0; the slot one position \emph{past} each assistant \seg{} is exactly $0$. Stage 1 trains for 8 epochs at base LR $1\times 10^{-4}$ on the Bridge and mask head; Stage 2 unfreezes \qwen{} LoRA at $1\times 10^{-6}$ and the masked \seg{} rows at $1\times 10^{-5}$, decaying the Bridge LR by $10\times$ to avoid co-adaptation churn.

Table~\ref{tab:diagnostics} gathers the quantitative signals that motivated the final design.

\begin{table}[htb]
\centering
\small
\resizebox{\linewidth}{!}{%
\begin{tabular}{llcc}
\toprule
Failure mode & Measurement & Value & Fix \\
\midrule
Context Leakage & \seg{} cosine, shared vs. isolated & 0.74 vs. 0.16 & block-diagonal extraction \\
Slot-1 Positional Bias & slot allocation on RWTD & 90.5\% / 0.0\% & batch multiplexing \\
Count Collapse & outputs with $K{=}1$ & 85\% & non-uniform LM weighting \\
Directional Drift & RWTD cosine $\Delta$ vs. ADE20K $\Delta$ & 0.183 vs. 0.048 & frozen-resizer Bridge \\
Semantic bottleneck & zero-shot \samthree{} RWTD mIoU & 0.928 & native-width alignment \\
\bottomrule
\end{tabular}
}
\caption{\textbf{Diagnostics behind the design.} Each mechanism targets a specific observed failure rather than adding capacity alone. The zero-shot \samthree{} number is a diagnostic upper bound for SAM's native text pathway.}
\label{tab:diagnostics}
\end{table}

\section{Constructing TextureADE}
\label{app:constructing_textureade}

This section expands the concise main-text description of TextureADE construction in Section~\ref{sec:data_eval}. We first state the score in closed form, then define each component as a deterministic function of the image's dense semantic mask. Throughout, $H$ and $W$ are image height and width, $|I|=HW$ is total pixel count, and a TEXTURE class is any ADE20K class outside the object-centric set used by the miner's class taxonomy.

\textbf{Total review score.} For an input image, the Detecture Miner computes
\begin{equation}
s \;=\; 0.20\,s_A \;+\; 0.60\,s_{\mathrm{geom}} \;+\; \mathrm{bonuses} \;-\; \mathrm{penalties},
\quad s\in[0,100],
\label{eq:miner_total}
\end{equation}
and accepts the image iff $s\geq 65$. Bonuses (additive, capped) reward high texture coverage and a region count in $\{2,3,4\}$; penalties (subtractive) downweight object-dominated, fragmented, mosaic, or low-texture-content scenes. Both are deterministic functions of the dense mask and auxiliary VLM tags; their exact weights and the tag set are recorded in the released configuration.

\textbf{Mask-structure score $s_A$.} Let $N$ be the number of raw dense-mask components in the image, let $\{a_i\}_{i=1}^{N}$ be their pixel areas with ratios $r_i = a_i/|I|$, let $r_{\max}=\max_i r_i$, $r_{\mathrm{med}}=\mathrm{median}_i\,r_i$, and let $\phi_{\mathrm{small}} = \tfrac{1}{N}\sum_i \mathbf{1}[r_i<10^{-3}]$ be the fraction of very small components. The Shannon entropy of the area distribution (base 2) is
\begin{equation}
H(\{r_i\}) \;=\; -\sum_{i=1}^{N} \tilde r_i\,\log_2 \tilde r_i,
\qquad
\tilde r_i = \frac{r_i}{\sum_j r_j}.
\label{eq:miner_entropy}
\end{equation}
Five normalized component scores are then defined, each clipped to $[0,1]$:
\begin{align}
n_{\mathrm{score}} &= \min\!\big(1,\, N/120\big), &
\ell_{\mathrm{score}} &= 1 - \min\!\big(1,\, r_{\max}/0.65\big), \nonumber\\
m_{\mathrm{score}} &= 1 - \min\!\big(1,\, r_{\mathrm{med}}/0.03\big), &
p_{\mathrm{score}} &= \min\!\big(1,\, \phi_{\mathrm{small}}/0.55\big), \\[-2pt]
e_{\mathrm{score}} &= \min\!\big(1,\, H(\{r_i\})/4.0\big). \nonumber
\end{align}
The mask-structure score is their convex combination, expressed on a $0$--$100$ scale:
\begin{equation}
s_A \;=\; 100 \cdot \big(0.28\,n_{\mathrm{score}} + 0.24\,\ell_{\mathrm{score}} + 0.18\,m_{\mathrm{score}} + 0.18\,p_{\mathrm{score}} + 0.12\,e_{\mathrm{score}}\big).
\label{eq:miner_sA}
\end{equation}
Intuitively, $n_{\mathrm{score}}$ rewards rich mask inventory, $\ell_{\mathrm{score}}$ penalizes a single dominating region, $m_{\mathrm{score}}$ penalizes a coarse median, $p_{\mathrm{score}}$ rewards a healthy population of small components, and $e_{\mathrm{score}}$ rewards an entropically balanced area distribution.

\textbf{Texture-boundary geometry score $s_{\mathrm{geom}}$.} Let $\mathcal{R}=\{R_1,\ldots,R_M\}$ be the connected components obtained by 4-connectivity labeling on TEXTURE-class pixels, and let $\mathcal{L}=\{R_j : a(R_j) \geq 0.08\,|I|\}$ be the subset of \emph{large} components. Define:
\begin{itemize}[leftmargin=1.2em, itemsep=0.15em, topsep=0.20em]
  \item $C_{\mathrm{large}} = |\mathcal{L}|$ -- the count of large textured regions.
  \item $C_{\mathrm{strong}} = \big|\{(R_i,R_j)\in\mathcal{L}\times\mathcal{L},\, i<j : |\partial(R_i,R_j)| \geq 40\}\big|$ -- the number of large-region pairs whose shared boundary $\partial(R_i,R_j)$ is at least 40 pixels long.
  \item $\beta = \tfrac{1}{2(H+W)}\sum_{i<j} |\partial(R_i,R_j)|$ -- total inter-region boundary length normalized by image perimeter.
  \item $b = \max\!\big(0,\, 1 - \tfrac{\max_{R_j\in\mathcal{L}} a(R_j)}{\sum_{R_k\in\mathcal{L}} a(R_k)}\big)$ -- area balance over large regions, in $[0,1]$.
  \item $\phi_{\mathrm{obj}} \in [0,1]$ -- fraction of pixels labeled as object-centric (people, vehicles, animals, products), used as a penalty multiplier.
\end{itemize}
The four positive terms are clipped to $[0,1]$ and combined:
\begin{align}
\rho_{\mathrm{score}} &= \min\!\big(1,\, C_{\mathrm{large}}/3\big), &
\sigma_{\mathrm{score}} &= \min\!\big(1,\, C_{\mathrm{strong}}/2\big), \nonumber\\
\beta_{\mathrm{score}} &= \min\!\big(1,\, \beta/0.35\big), &
b_{\mathrm{score}} &= b,
\end{align}
\begin{equation}
s_{\mathrm{geom}}^{\mathrm{raw}} \;=\; 100 \cdot \big(0.28\,\rho_{\mathrm{score}} + 0.34\,\sigma_{\mathrm{score}} + 0.26\,\beta_{\mathrm{score}} + 0.12\,b_{\mathrm{score}}\big),
\label{eq:miner_sgeom_raw}
\end{equation}
\begin{equation}
s_{\mathrm{geom}} \;=\; s_{\mathrm{geom}}^{\mathrm{raw}}\cdot\big(1 - \min(0.75,\, 1.15\,\phi_{\mathrm{obj}})\big).
\label{eq:miner_sgeom}
\end{equation}
Equation~\eqref{eq:miner_sgeom_raw} rewards a target of three large textured regions ($\rho$), at least two strong inter-region boundaries ($\sigma$), substantial total boundary length relative to image perimeter (boundary coherence $\beta$), and balanced large-region areas ($b$). Equation~\eqref{eq:miner_sgeom} damps the geometry score on object-dominated scenes; the cap of $0.75$ prevents fully zeroing out images that contain a small object alongside genuine texture structure.

\textbf{Acceptance rule and post-acceptance merging.} An image is accepted into the curated corpus iff $s\geq 65$. For accepted images, per-class TEXTURE masks are merged into at most five regions, each covering at least $1\%$ of image area; smaller fragments are absorbed into the nearest large neighbor or relabeled as background.

\textbf{Dual-source mining: training vs. evaluation.}
The same scoring pipeline is run on two distinct corpora. (i)~For the \emph{training} set, the Miner is applied to TextureSAM's CNT-augmented~\cite{cnt} ADE20K \emph{training} split (a pool of order $10^5$ candidate images) and accepts only those scoring ${\ge}65/100$, yielding a curated set of ${\sim}14{,}000$ multi-texture samples. The mining is therefore a strict \emph{filter} over augmented data. (ii)~For the \emph{evaluation} set, the identical Miner is applied directly to the \emph{natural} ADE20K \emph{validation} split, accepting 212 samples by the same threshold; this gives \emph{TextureADE}. Test data is unaugmented natural imagery; training data is aggressively filtered augmented imagery. Mask-derived geometry is identical across the two paths -- only the source corpus differs.

Claude Haiku~4.5 (model id \texttt{claude-haiku-4-5-20251001}) annotations use the fixed pattern ``Texture of <visual features>, <spatial context>'' and target short descriptions of roughly 10--15 words. Claude is used here as a language-labeling tool, not as a segmentation model. The annotation is intentionally tied to a mask overlay rather than to free-form image captioning: the VLM describes a proposed region, it does not invent the region. The resulting corpus contains ${\sim}14{,}000$ training samples and an average of 3.2 texture regions per image.

Throughout the paper, Detecture refers to the model, Detecture Miner to the curation pipeline, and TextureADE to the ADE20K-derived dataset contribution produced by that pipeline. Earlier drafts used nearby Detecture-based dataset names; here we use TextureADE consistently to avoid confusion between the model and the dataset.

The ADE20K validation split is processed with the same miner to produce TextureADE, the ADE20K-derived benchmark route used in the main paper.

\textbf{Why TextureADE: scarcity, evidentiary split, and the self-reference concern.}
Texture segmentation has a data bottleneck. RWTD is the strongest real-world benchmark available, but with $253$ images it is too small to support training or broad ablation on its own. Other resources -- DTD, KTH-TIPS, MINC, OpenSurfaces, dense material segmentation datasets -- are organized around categories or surface labels rather than blind multi-region texture partitioning~\cite{dtd,sharan2013material,minc,opensurfaces,densematerialseg}. TextureADE closes that gap by mining a texture-specific route from ADE20K, a canonical dense-mask scene-parsing corpus already used widely in segmentation research. The same mined corpus also provides the training data required to learn the bridge from sub-semantic language to masks.

The setup can look self-referential, because the training corpus is ADE20K-derived and one evaluation route is ADE20K-derived as well. We separate the evidentiary roles explicitly: TextureADE is the in-domain but held-out realistic-scale route, asking whether a method can train and operate in blind variable-$K$ mode at all. RWTD, distributionally independent and small enough to inspect, carries the cross-domain stress test -- whether the learned language-to-mask interface transfers beyond the ADE20K-derived mining regime. As TextureSAM already showed by building its own texture supervision from ADE20K~\cite{texturesam}, the alternative is usually to choose between tiny real benchmarks and synthetic data; in that sense, mining TextureADE from a well-known dense-mask corpus is part of the contribution itself. Several failed variants during development looked stable on TextureADE while degrading on RWTD, which is exactly why the paper treats the two as complementary rather than substitutes.

The division of labor is deliberate: Detecture Miner constructs candidate regions and descriptions; \method{} learns to discover and ground them at inference time. The scientific claim is therefore not that the miner itself performs the task, but that once realistic supervision exists, sub-semantic language grounding can be trained without collapsing the partitioning problem.

\section{Ablation Trail}
\label{app:ablation}

This section expands the brief ablation summary in Section~\ref{sec:experiments}. Table~\ref{tab:ablation} traces the path from the failed baseline to the final live model. The rows mix two diagnostic regimes. In teacher-forced rows, the correct region descriptions are supplied so that only the mask pathway is being tested. In live rows, the model must emit its own descriptions and grounding tokens, so the full system is under test. The clearest lesson is that teacher-forced mask quality and live inference can diverge: the naive exponential cliff improved oracle geometry but produced 0.000 live RWTD mIoU because the model stopped emitting \seg. Shifted-Zero plus masked-row \seg{} training closes that gap, and the final checkpoint reaches 0.8053 RWTD mIoU in the benchmark suite.

Several ablation rows are intentionally diagnostic rather than final-system comparisons. The teacher-forced rows bypass generation to test whether the geometry path is healthy. The zero-shot \samthree{} row bypasses the VLM and projector to test the semantic capacity already present in SAM's native text pathway. These rows are valuable because they separate failure causes: a poor teacher-forced score indicates a geometric or projector problem; a good teacher-forced score paired with poor live inference indicates a language-emission problem; and a strong native-SAM score paired with a weak projected score indicates that the projector is destroying information.

\begin{table}[htb]
\centering
\small
\resizebox{\linewidth}{!}{\begin{tabular}{p{0.49\linewidth}cc}
\toprule
Configuration & RWTD mIoU & Interpretation \\
\midrule
High-capacity projector, causal multi-\texttt{[SEG]}, full LM loss & 0.544 & baseline failures remain \\
+ batch-multiplexed SAM decoding & 0.703 & removes Slot-1 positional bias \\
+ slim $4096{\to}512{\to}256$ bottleneck projector & 0.732 & reduces directional drift \\
+ block-diagonal extraction, teacher-forced & 0.810 & removes context leakage \\
Same family, live generation without LM supervision & 0.136 & language collapse exposed \\
+ cosine-decayed LM weighting, live exactly-$2$ & 0.694 & partial recovery \\
Bridge-to-Native-Space + naive $\lambda(\texttt{[SEG]})=0$ cliff, live & 0.000 & emission failure exposed \\
+ Shifted-Zero + masked-row \texttt{[SEG]} warm-start, live epoch 16 & 0.794 & emission recovered \\
Final checkpoint used in benchmark & \textbf{0.8162} & final live model \\
SAM~3 native text prompt, no training & 0.928 & diagnostic upper bound \\
\bottomrule
\end{tabular}
}
\caption{\textbf{RWTD ablation trail.} The zero-shot \samthree{} row is a diagnostic upper bound for SAM's native text pathway, not an equally autonomous texture model.}
\label{tab:ablation}
\end{table}

\subsection{Semantic Anchor Ablation -- Visual Evidence}
\label{app:semantic_anchor_ablation}

This subsection provides visual support for the strict head-noun
omission ablation summarised in Section~\ref{sec:experiments} and
quantified in Table~\ref{tab:semantic_ablation}. The ablation
deterministically strips the leading head noun (e.g.,
``\textit{quilt}'') from each \texttt{TEXTURE\_k} description at
inference time and recomputes the \seg{} hidden state, isolating the
contribution of the appearance modifiers (material, micro-pattern,
color, spatial role) to the predicted partition.

\begin{figure*}[!t]
\centering
\includegraphics[width=\linewidth]{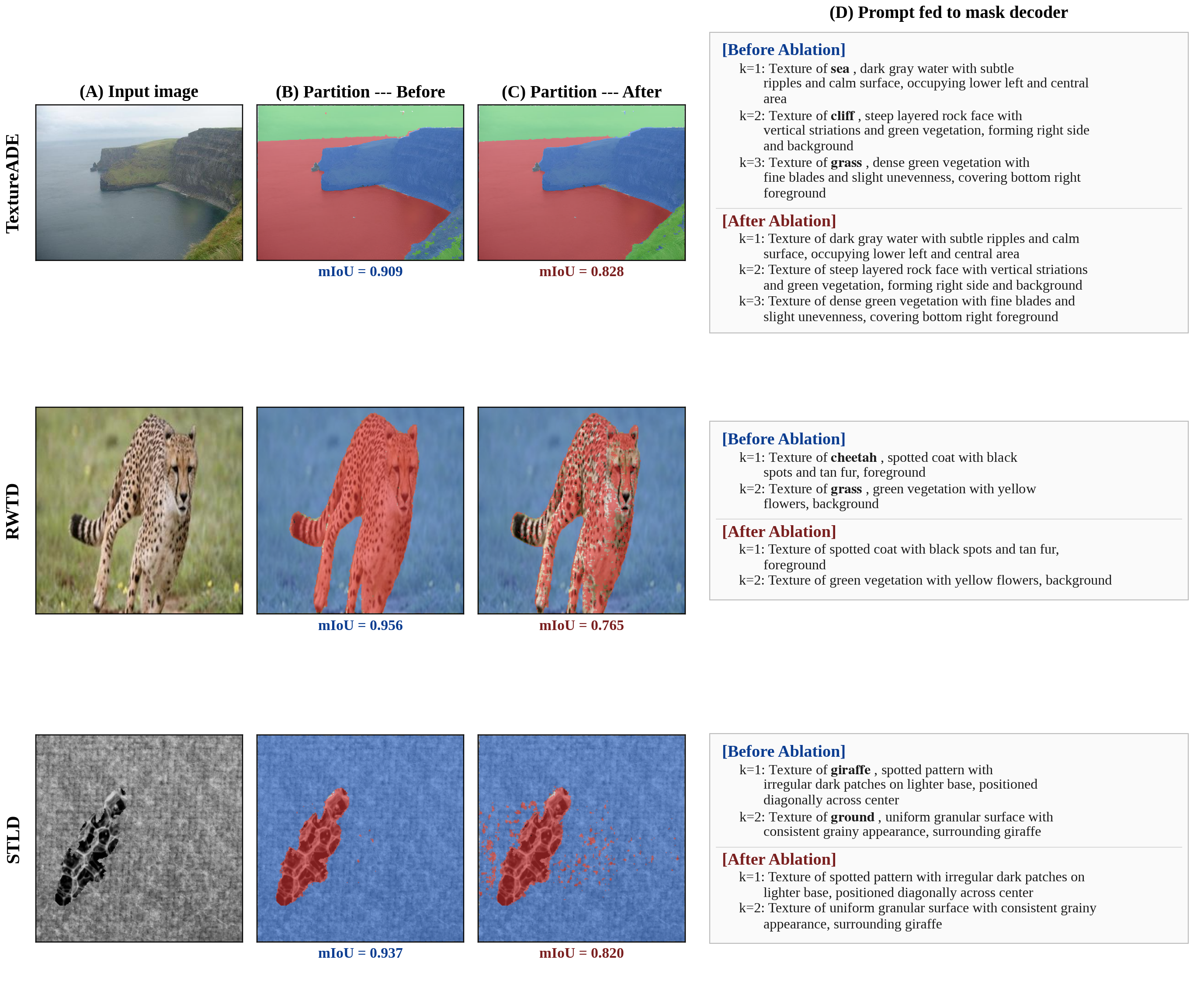}
\caption{\textbf{Semantic Anchor Ablation -- visual evidence
accompanying Table~\ref{tab:semantic_ablation}.} Three representative
examples (one per evaluation route: TextureADE, RWTD, STLD). Each row
contains four panels: \textbf{(A) Input image}; \textbf{(B)} \method{}
predicted partition under the \emph{Before-Ablation} prompt;
\textbf{(C)} \method{} predicted partition under the
\emph{After-Ablation} prompt; \textbf{(D)} the two prompts fed to the
mask decoder, with the head noun bolded in red in the
\textbf{[Before Ablation]} block (e.g., ``\textit{Texture of
\textbf{cheetah}, spotted coat with black spots and tan fur,
foreground}'') and removed in the \textbf{[After Ablation]} block
(``\textit{Texture of spotted coat with black spots and tan fur,
foreground}''). Per-cell mIoU values are reported beneath the partition
panels. Across all three benchmarks the After-Ablation mask remains a
coherent partition aligned with the same texture regions, confirming
that the \seg{} hidden state encodes appearance evidence rather than
the class identity that the head noun would supply -- the central
visual illustration of the paper's sub-semantic claim.}
\label{fig:semantic_anchor_ablation}
\end{figure*}

\textbf{Qualitative nuances of anchor omission.} Visual inspection of the ablation reveals two distinct behaviors. First, for inherently object-centric patterns (e.g., the cheetah in RWTD or the giraffe-pattern in STLD), omitting the explicit noun causes a slight expected degradation, as the semantic category provides a strong natural prior. Second, and more remarkably, the pure sub-semantic prompts can sometimes yield finer partitions than the dataset's own ground truth. In the TextureADE example (Figure~\ref{fig:semantic_anchor_ablation}, top row), the ADE20K-derived ground truth imprecisely merges the rocky cliff face and the overlying vegetation into a single coarse semantic ``cliff'' mask. The purely sub-semantic prompt successfully disentangles the rock from the grass based on appearance. Ironically, this superior textural precision registers as a drop in mIoU ($0.909 \rightarrow 0.828$) simply because the model is penalized for disagreeing with a coarse semantic label.

\section{Evaluation Protocol Details}
\label{app:eval_protocol}

The main paper uses three benchmark routes because they stress different parts of the claim; this appendix records the exact evaluation protocol. RWTD and STLD use a controlled exact-$K$ two-texture setting. This is not an advantage for \method{}, but the task definition for those routes: each image is evaluated as a two-region texture separation problem, analogous to binary foreground/background segmentation where the number of regions is fixed by the benchmark. \method{} and Sa2VA therefore receive an exact-two instruction. Because the baseline systems -- including both SAM-family models and single-target reasoning VLMs like Sa2VA -- lack native competitive multi-partition logic or autonomous stopping rules, the benchmark harness constructs a two-region partition for them (e.g., via the standard inverse-mask rule for SAM-based models). As established in Section~\ref{sec:data_eval}, this is a generous concession that artificially solves the count and partitioning problem, allowing these baselines to be evaluated purely on boundary quality without suffering fragmentation penalties.

TextureADE covers the complementary blind variable-$K$ regime. It is the in-domain but held-out realistic-scale route: images contain more than two texture regions, \method{}, Sa2VA, and Grounding-SAM~3 are not given the ground-truth $K$, and each method must operate in blind variable-$K$ mode. The only exception is SAM~3, for which we allow a top-$K_{\mathrm{GT}}$ proposal concession because its native proposal route has no autonomous stopping rule. This concession favors the baseline rather than \method{}: it gives SAM~3 the hidden count while \method{} must infer its own output set.

\textbf{Per-baseline instantiation.} Each baseline receives the same image and the route's count instruction, and is invoked through the interface for which it was designed. \samthree{}~\citep{sam3} (vanilla) is invoked through its native promptable path: on RWTD/STLD it receives an empty text prompt and the inverse-mask construction supplies the second region; on TextureADE it is run in dense-proposal mode and the top-$K_{\mathrm{GT}}$ concession (the only baseline-favoring concession in the suite) keeps the highest-scored proposals. \emph{Grounding-\samthree{}} (Grounding-DINO~\cite{groundingdino}~$\to$~\samthree{}) receives a fixed open-vocabulary detection prompt; on RWTD/STLD the inverse-mask rule applies, and on TextureADE its detector emits its own count without a $K_{\mathrm{GT}}$ concession. \emph{Sa2VA}~\cite{sa2va} receives a single conversational instruction (``\textit{describe and segment all distinct textures; output exactly $K=2$ regions}'' on RWTD/STLD, the open-range variant on TextureADE) and is decoded in single-pass mode, matching its intended single-target referring regime. \emph{TextureSAM}~\cite{texturesam} (a SAM-2 checkpoint fine-tuned on CNT-augmented ADE20K) is invoked through its native promptable path with the inverse-mask rule on RWTD/STLD; on TextureADE it is run with a uniform grid of foreground prompts and the inverse-mask completion of each, after which the partition harness performs Hungarian alignment against the ground truth. \method{} receives the exact-$K{=}2$ instruction on RWTD/STLD and the open-range instruction on TextureADE; in both cases its decoded \seg{} states drive the same Winner-Takes-All partition path described in Section~\ref{sec:method}.

Predictions are scored through the same partition path: logits are resized, a dustbin/background channel is included, and Hungarian matching is used to align predicted and ground-truth regions for mIoU. ARI is computed over the full partition and is therefore not equivalent to mIoU. The paper therefore reports metric-specific conclusions rather than collapsing mIoU and ARI into a single summary claim.

\textbf{Standard SAM~3 Benchmarks and Zero-Shot Preservation.} We deliberately evaluate on texture-specific datasets (RWTD, STLD, TextureADE) rather than standard SAM~3 benchmarks (e.g., COCO, LVIS, or COCO-Stuff). Standard benchmarks primarily evaluate object-centric boundaries or fixed-vocabulary ``stuff'' categories, which fundamentally differ from free-form, sub-semantic texture partitioning. Evaluating on them would test semantic object labels rather than the autonomous discovery of visual regularities. Furthermore, because \method{} keeps the entire \samthree{} backbone strictly frozen, we guarantee that its native zero-shot and promptable segmentation capabilities remain completely intact and undisrupted.


\section{Full Autonomous Texture Descriptions}
\label{app:full_descriptions}

\makeatletter
\setlength{\@fptop}{0pt}
\setlength{\@fpsep}{12pt plus 0fil}
\setlength{\@fpbot}{0pt plus 1fil}
\makeatother

This appendix reproduces fifteen novel samples (five per evaluation
route, disjoint from main-paper figures) with their complete
assistant-region descriptions, exactly as \qwen{} emits them at
inference -- no editing, truncation, or template post-processing;
only the trailing \seg{} grounding token is removed for readability.

\begin{figure}[!t]
\centering
\includegraphics[width=\linewidth]{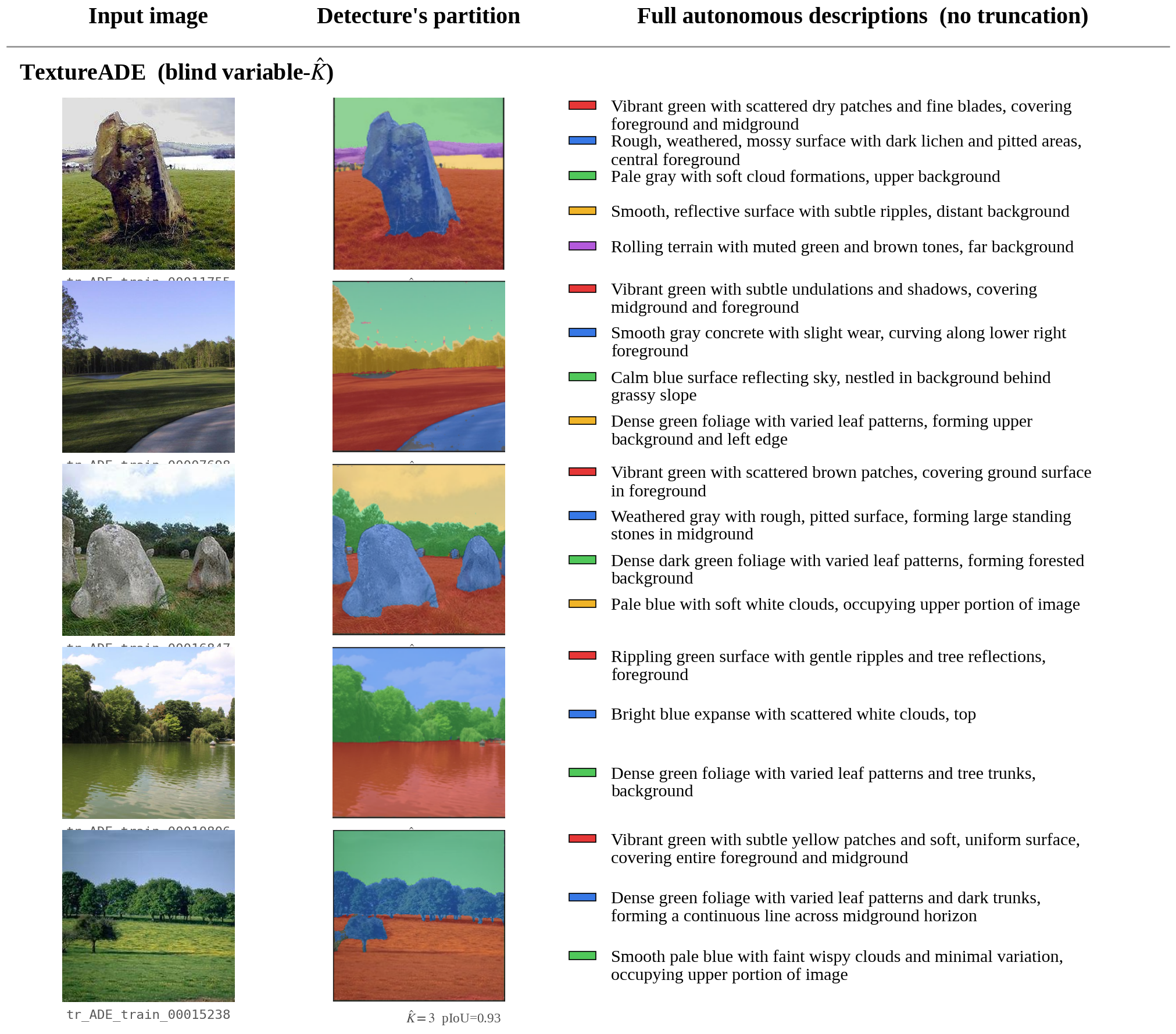}
\caption{\textbf{TextureADE -- five novel blind variable-$\hat{K}$
examples}. $\hat{K}\in\{3,4,5\}$). Each
row shows the input image, \method{}'s predicted partition, and the
complete free-form text emitted by \qwen{} for every \seg{} slot --
no length truncation or template post-processing. Color squares
match the partition palette so each phrase aligns to its mask.}
\label{fig:full_descriptions_textureade}
\end{figure}

\begin{figure}[!t]
\centering
\includegraphics[width=\linewidth]{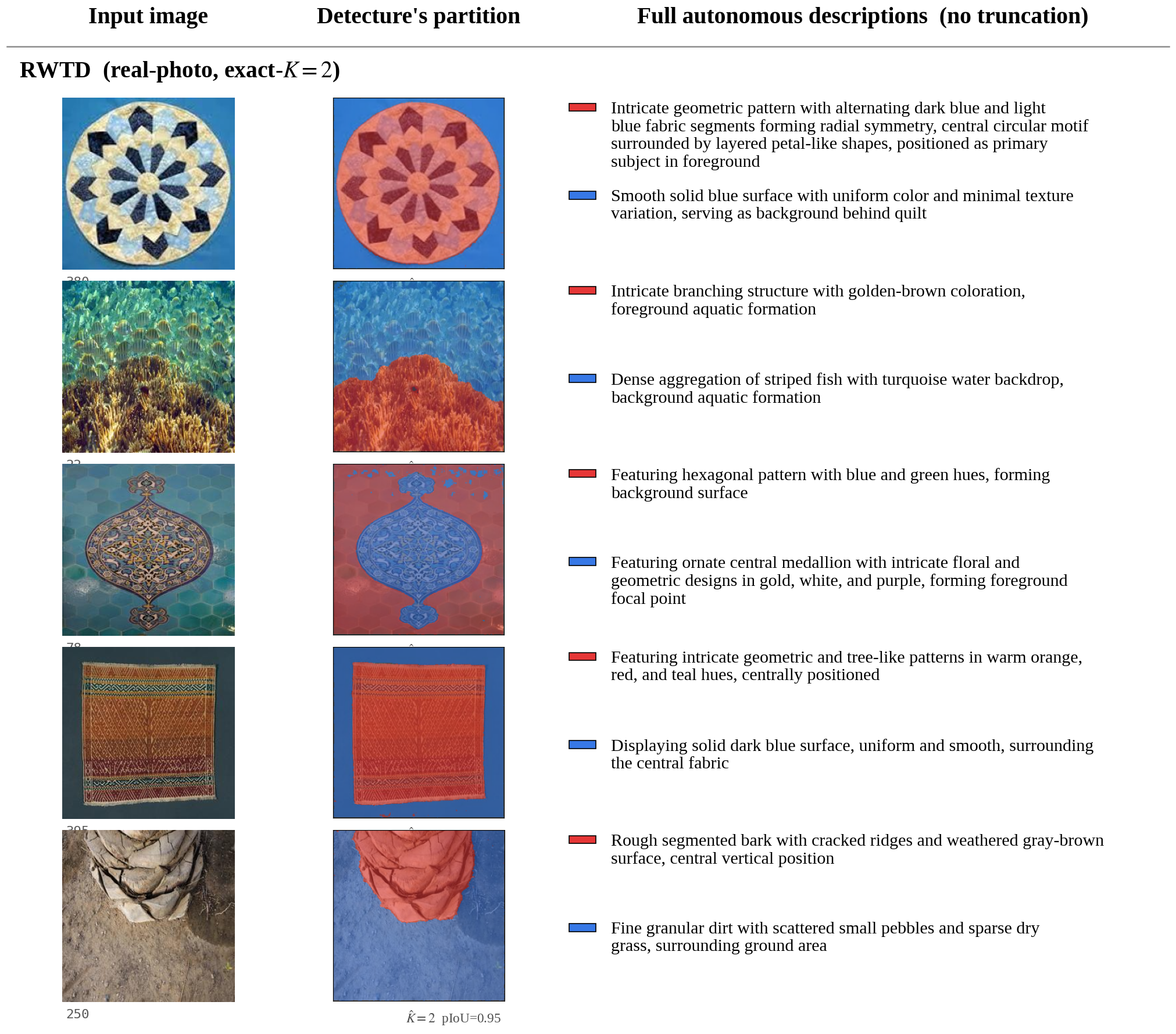}
\caption{\textbf{RWTD -- five novel real-photo exact-$K{=}2$
examples}. The descriptions cover
diverse domains -- intricate quilting, ornate Mediterranean
tilework, patterned textile, biological scenes (coral reef with
striped fish school, palm-bark macro) -- demonstrating that the
sub-semantic grounding generalizes beyond the ADE20K-derived
training distribution.}
\label{fig:full_descriptions_rwtd}
\end{figure}

\begin{figure}[!t]
\centering
\includegraphics[width=\linewidth]{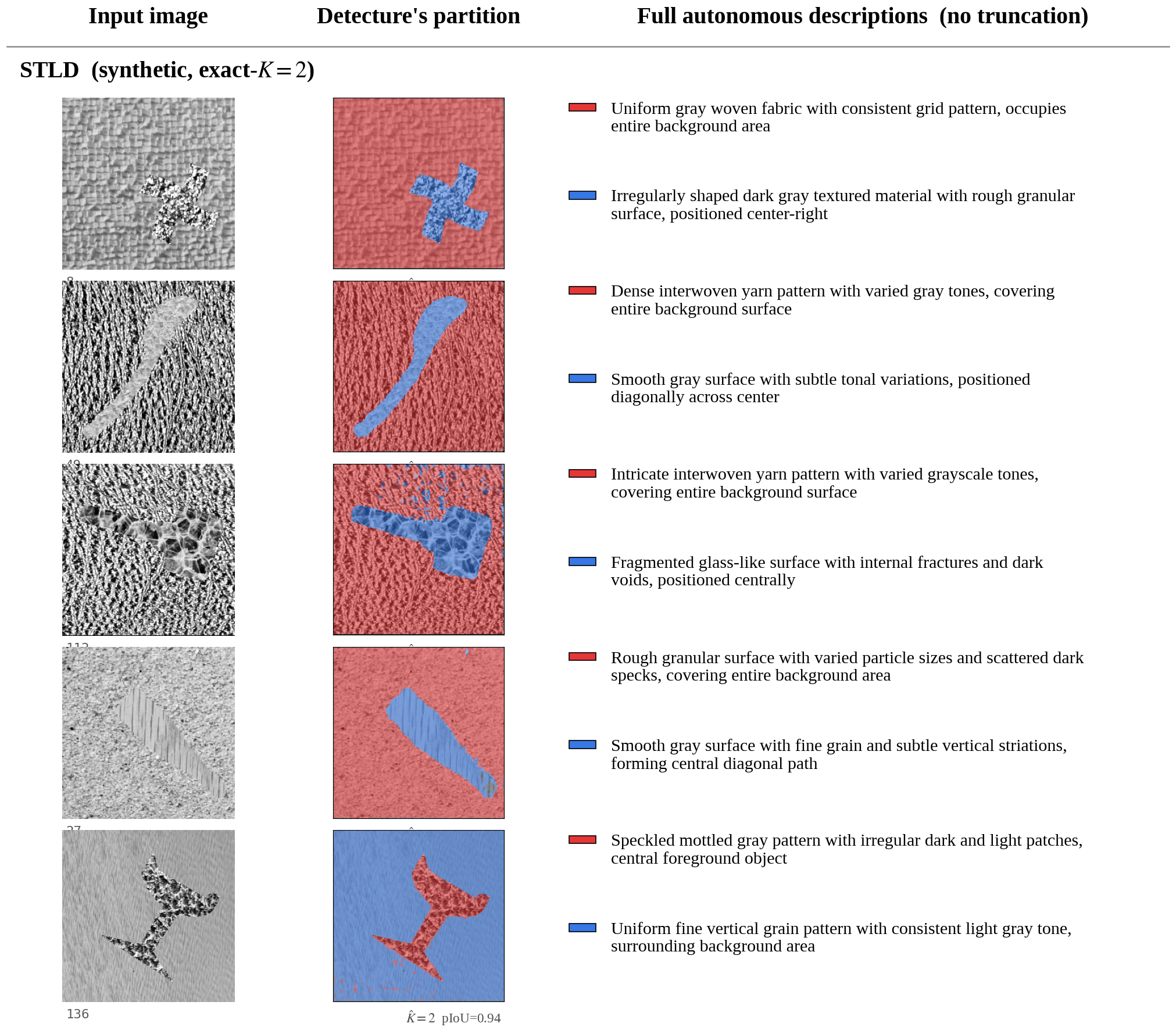}
\caption{\textbf{STLD -- five novel synthetic exact-$K{=}2$
examples}. Knitted-fabric backgrounds
with embedded fragmented or curved foreground objects, gravel /
concrete contrasts, and an airplane-on-sky composite. The phrases
remain appearance-anchored (head category, two or three appearance
attributes, spatial qualifier) even on synthetically composited
scenes, where standard category language would not exist.}
\label{fig:full_descriptions_stld}
\end{figure}

\textbf{Selection rationale.} The fifteen samples expose the axis of
variation each evaluation route tests.
\emph{TextureADE} (Figure~\ref{fig:full_descriptions_textureade}):
the five rows span the blind variable-$\hat{K}$ range -- one
$\hat{K}{=}5$ landscape, two $\hat{K}{=}4$ scenes mixing natural and
human-made surfaces, and two $\hat{K}{=}3$ scenes anchoring the
high-pIoU regime.
\emph{RWTD} (Figure~\ref{fig:full_descriptions_rwtd}): real-photo
domain breadth at exact-$K{=}2$ -- decorative quilting, underwater
coral with a striped-fish school, ornate Mediterranean tilework
where both regions share the head noun ``tile'' and must be
separated on appearance alone, patterned textile against a solid
ground, and a palm-bark / soil macro.
\emph{STLD} (Figure~\ref{fig:full_descriptions_stld}): the
synthetic-composite difficulty cases -- an opaque granular blob on
woven fabric, a smooth elongated shape on a knitted background, a
fragmented glass-like translucent object with highly irregular
boundaries, gravel / concrete juxtaposition, and an airplane on a
noise-grain ``sky.'' Every sample ID is verified disjoint from
Figures~\ref{fig:teaser} and~\ref{fig:qualitative_results}.

\textbf{Structural signature.} Across all three routes the
descriptions exhibit the same structural signature the architecture
is designed to elicit: a head category, two or three appearance
attributes (color, micro-structure, surface state), and a spatial
qualifier. This is the same triple the training prompt requests,
emitted \emph{autonomously} on every held-out sample -- evidence
that the \seg{} hidden state is grounded in appearance rather than
in a learned vocabulary of category labels. While the VLM naturally
emits this semantic head noun, the strict ablation in
Section~\ref{sec:experiments} proves it is computationally
redundant: the \seg{} hidden state is grounded entirely in the
sub-semantic appearance attributes rather than the category label.

\FloatBarrier

\FloatBarrier

\section{Linguistic Richness of Descriptions}
\label{app:linguistic_richness}

\begin{figure}[!htbp]
\centering
\includegraphics[width=0.85\linewidth]{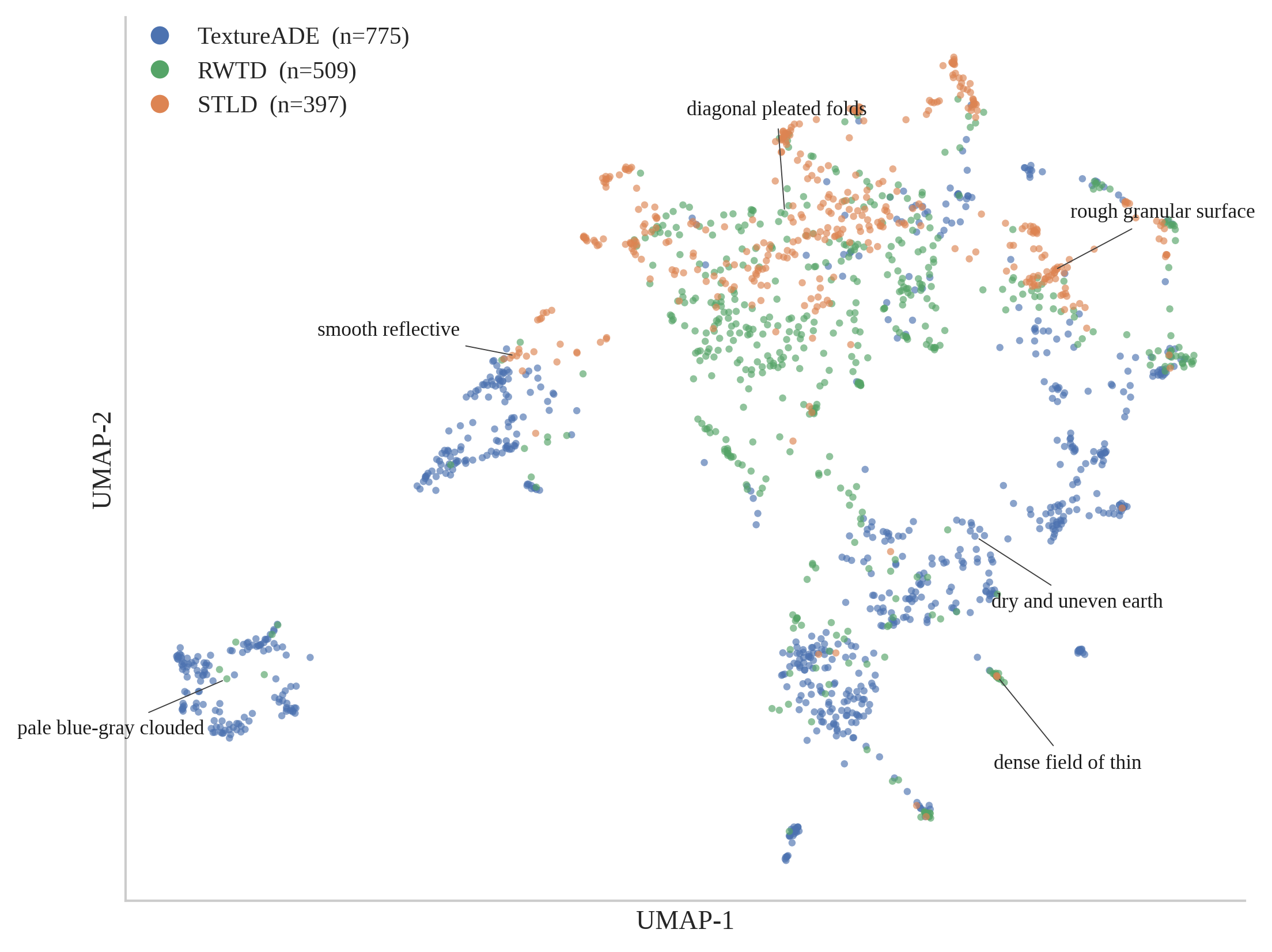}
\caption{\textbf{Semantic landscape of \method{}'s sub-semantic
descriptions.} 2D UMAP projection of \texttt{all-MiniLM-L6-v2}
sentence embeddings of every \texttt{TEXTURE\_i} description
\method{} emits across the three benchmarks ($n{=}1{,}681$, canonical
``Texture of $\langle\text{body}\rangle$'' inference form; cosine
metric, $n_{\mathrm{neighbors}}{=}25$,
$\min\_\mathrm{dist}{=}0.15$). Points are coloured by source
benchmark.  The descriptions form a single continuous manifold with
substantial cross-benchmark overlap: \method{} produces a shared
sub-semantic vocabulary, not three dataset-specific dialects.  Six
representative descriptions are annotated to illustrate the breadth
of the texture vocabulary across the cloud.}
\label{fig:semantic_diversity_umap}
\end{figure}

\textbf{Population-level view of \method{}'s sub-semantic descriptions.}
A natural concern about any VLM-driven segmentation system is that
the language head collapses onto a small set of rigid templates and the
``descriptions'' carry little real textural content.  To rule this out
for \method{}, we embed every \texttt{TEXTURE\_i} description emitted
across all three benchmarks ($n{=}1{,}681$ in the canonical
``Texture of $\langle\text{body}\rangle$'' inference form) with the
sentence-transformer \texttt{all-MiniLM-L6-v2} and project the
$384$-dimensional embeddings to two dimensions with UMAP under a
cosine metric ($n_{\mathrm{neighbors}}{=}25$,
$\min\_\mathrm{dist}{=}0.15$).  Figure~\ref{fig:semantic_diversity_umap}
shows the resulting 2D landscape, coloured by source benchmark.
Three properties stand out.  First, the descriptions form a single
continuous manifold rather than a handful of point-like clusters --
there is no template collapse.  Second, the three benchmarks occupy
heavily overlapping regions of the manifold, indicating that
\method{} produces a \emph{shared sub-semantic vocabulary} rather than
dataset-specific phrasing.  Third, we annotate six representative
samples spread across the manifold so the reader can verify the
spread directly: every annotation describes material/structure
(\emph{rough granular surface}, \emph{diagonal pleated folds},
\emph{dense field of thin elongated blades}, \ldots) rather than
naming an object, illustrating the breadth of the texture
vocabulary \method{} produces.

\end{document}